\documentclass[journal]{IEEEtran}
\usepackage{amsmath}
\usepackage{textcomp}
\usepackage{hyperref}
\usepackage{float}
\usepackage[caption = false]{subfig}
\usepackage{graphicx}

\ifCLASSINFOpdf
\else
\fi
\begin{document}
%
\title{Particle Swarm Optimization Based Demand
Response Using Artificial Neural Network Based
Load Prediction}
%
%
%

\author{Nasrin~Bayat
\thanks{N.Bayat is with the Department
of Electrical and Computer Engineering, University of Central Florida, Orlando,
FL, 32816 USA e-mail: (nasrinbayat@knights.ucf.edu).}}

%
%

\markboth{Journal of \LaTeX\ Class Files,~Vol.~14, No.~8, August~2015}%
{Shell \MakeLowercase{\textit{et al.}}: Particle Swarm Optimization Based Demand
Response Using Artificial Neural Network Based
Load Prediction}
%



\maketitle

\begin{abstract}
In the present study, a Particle Swarm Optimization (PSO) based Demand Response (DR) model, using Artificial Neural Network (ANN) to predict load is proposed. The electrical load and climatological data of a residential area in Austin city in Texas are used as the inputs of the ANN. Then, the outcomes with the day-ahead prices data are used to solve the load shifting and cost reduction problem. According to the results, the proposed model has the ability to decrease payment costs and peak load.
\end{abstract}

\begin{IEEEkeywords}
demand response, differential evolution, artificial neural network, load forecasting, particle swarm optimization, smart grid.
\end{IEEEkeywords}

%
\IEEEpeerreviewmaketitle

\section{Introduction}
%
%
%
%
\IEEEPARstart{I}{n} the recent decades, we have been facing with energy and environmental crisis, which have caused problems such as emptying natural resources, and the green house effect. Smart grids are considered as a solution for these problems \cite{kerdphol2016optimum}. One of the most important goals in smart grid is energy efficiency \cite{6861959}. Therefore, in order to enhance the efficiency, new technologies, such as demand response (DR) are incorporated into smart grid. The DR program can persuade end-user customers change their electricity usage pattern, in response to incentives respecting the energy saving, electricity price, cost reduction, and optimization of the grid operation \cite{8681422}.
There are three major kinds of consumers that the DR programs are applied to them residential, commercial and industrial consumers. A residential DR is presented in \cite{haider2016residential}, which is based on adaptive consumption pricing and enables customers to lower their energy consumption and utilities to manage total load. The proposed model encourages consumers to have an active participation in the DR program by fitting energy costs to consumers' consumption levels. An incorporation of energy storage systems (ESS) that operate as loads, such as electric vehicle (EV) and uninterruptible power supply (UPS) with utility with the help of DR for minimizing the costs is presented in \cite{8385345}. The loads are planned with the help of particle swarm optimization (PSO) algorithm. PSO is an effective way of solving large-scale non linear optimization problems \cite{del2008particle}. PSO based DR methodology that contributes to making the load curve flatter is presented in \cite{kinhekar2015particle}. Load shifting DR technique is used to modify the load curve of the system. A PSO based methodology, is proposed in \cite{faria2013modified} the goal is to help a hypothetical power player that manages the resources in a distribution network and the network itself, in reducing operation costs. In the present work, PSO is suggested to solve load shifting DR management problem.

In addition, electrical load forecasting plays a vital role in achieving the concept of the next generation power system such as efficient energy management, smart grid and better power system planning. Ordinarily, load forecasting can be divided in three classes on the basis of time intervals, namely long, medium and short term load forecast. Short term load forecast plays a crucial role in efficient control of spinning reserve, unit commitment and evaluation of sales or purchase contracts between different companies \cite{raza2015review}. Recently, soft computing (SC) methods has been proved to be more effective in forecasting than traditional methods. Among SC methods, artificial neural network (ANN) is the most commonly used forecasting method \cite{chang2011monthly}. ANN can be defined as an array of fundamental processors called neurons, that are highly connected \cite{baliyan2015review}, and a perfect depiction of it is presented in \cite{simon2004comprehensive}, \cite{1527779}.
ANNs have been applied to a plethora of areas of statistics such as time series prediction \cite{lee2009time}, \cite{crone2011advances}, and \cite{gheyas2011novel}. In this paper a time series prediction method and the multi layer feed forward ANN model, that can learn complicated and non-linear relationships, is employed to predict short-term load based on electric power consumption and climatological data of a residential area in Austin city.
After this introduction section, section II explains the problem formulation and the detailed methodology. Simulation results are presented in section III, and conclusions are given in section IV.

\section{Problem Formulation and Methodology}
\subsection{Detailed Methodology}
First, the hourly electric power consumption data of a residential area in Austin, is obtained from the Electric Reliability Council of Texas (ERCOT) website \cite{Electric}, and the corresponding local climatological data is extracted from National Centres for Environmental Information website \cite{Electric1} Furthermore, Day-ahead hourly prices are obtained from \cite{Electric3}. Data pre-processing and feature selection play an important rule in machine learning. There are different methods for pre-processing and extracting features from dataset \cite{gupta2021critical}, \cite{hakak2021ensemble}. Bag-of-words feature representation method is used in \cite{bayat2022human} and it is proved that its performance is better than statistical feature extraction methods. In bag-of-words method dataset is divided in to sliding windows and similar sequences of data are given the same word. In this way we will have a combination of words and we can classify them \cite{bayat2022human}. In this paper, pre-processing is done by normalizing ANN input data to [-1,1] range, to avoid convergence problems. The data is related to the year of 2010 . ANN is represented as a set of layers, which are divided into three categories namely, input, hidden, and output. in terms of the number of hidden layers and neurons in each layer, based on trying different cases and altering the number of neurons from 15 to 70 , the best number of hidden layers is determined to be 4 , with 25 neurons in the first layer, 20 in the second and 15 in the third layer and 1 in the output layer. If the number of neurons in the hidden layer is too big, the model can be overfitted with a small fluctuation in the data. On the other hand, if it is too small, the ANN can not predict properly. The type of training function is "TANSIG". The dataset during the first four months of 2010 is separated into two subsets. The first subset which starts from first of the January to April 6 , is the training subset, and the second one which starts from April 7 to April 30 , is the test subset.
Second, an ANN model is used to predict day-ahead electric load with the help of MATLAB. The inputs of ANN are hourly average wind speed, outside temperature, average heat index, average cold index, average dew point, and hourly power consumption of previous hours. The target is the 24 hour-ahead electrical demand.
Third, to solve the load shifting problem, PSO algorithm is evolved and tested in MATLAB. The problem formulation of this part is based on the DR load shifting strategy given in [18]. The PSO method is based on the mathematical model proposed in subsection B. The goal is to minimize load shifting and energy operation costs. The objective function includes two normalized terms, load shifting and cost. In order to provide a give-and-take between load shifting and total operating costs, weighting coefficients are used. They provide a set of solutions and based on the preferences and energy management capabilities, the best solution will be achieved. There are several factors that can affect the preferences. As an instant, during a specific period of time, the minimization of the load shifting can be more important than reduction of the costs.
In this paper, Differential Evolution (DE) algorithm is also used to solve the DR program and the outcomes are compared with the PSO results. PSO and DE are meta-heuristic algorithms. They have a simple model with few parameters and have acceptable convergence. Additionally, they are population-based and derivative-free. PSO and DE algorithms are explained in subsection C and D, respectively. Moreover, the objective function for both of them is as stated in (1). The inputs of both algorithms are day-ahead hourly prices and ANN based hourly predicted load.
\subsection{PSO-based DR methodology}
The proposed day-ahead PSO scheme which uses ANN hourly load predictions, is formulated as the minimization problem below. Two criteria, namely normalized cost and load shifting, form the objective function, as shown in (1) \cite{kampelis2018development}:
Minimize:
\begin{equation}
OF = w_1\times\frac{E_c}{E_{cmax}}+w_2\times \frac{L_{sh}}{L_{shmax}}+\text {alpha}\times\text { violation }
\end{equation}
Where, $w_1$ and $w_2$ are weighting coefficients, $E_c$ is hourly energy operation cost $(\$)$, $E_{cmax}$ denotes the normalization index of cost criterion ($\$$), $L_{sh}$ signifies total load shift in a day $(KWh)$, and $L_{shmax}$ is the normalization index of load shift criterion $(KWh)$. Alpha is violation coefficient. violation is determined with (2).
\begin{equation}
\text { violation }=\max \left(\left(\frac{L}{L_{pre}}\right)-1,0\right)
\end{equation}
Where $L$ is total load of a day $(KWh)$, and $L_{pre}$  is the predicted total load of that day $(KWh)$. If violation is equal to zero, there will not be any deviation between the predicted and the optimized solution. $E_c$ and $L_{sh}$  are computed with (3) and (4), respectively.
\begin{equation}
E_{c}=\sum_{h=1}^{24} L_{h} \times P_{h}
\end{equation}
\begin{equation}
L_{sh}=\sum_{h=1}^{24} \mid L_{h}-L_{hpre}\mid
\end{equation}
Where $L_h$ is hourly load $(KWh)$, $P_h$ is day-ahead hourly
price of energy $\frac{\mbox{\textcent}}{Kwh} $, and $L_{hpre}$ is predicted hourly load
$(KWh)$. Day-ahead hourly prices are obtained from [21].
\subsection{Particle Swarm Optimization Algorithm}
In the PSO algorithm, there are a number of candidate solutions, that are called particles scattered in the search space. Each particle determines the value of the objective function using its position information. Then, it chooses a direction to move, using the information of its current position, the best position that it had experienced, and the information of one or more particles of the best particles available in the population. After a collective movement, one step of the algorithm is finished. This step will be reiterated several times until the desired solution is obtained. Eq. 5 and 6 describe the behavior of the particles.
\begin{equation}
\begin{aligned}
&v_{t+1}=w \times v_{t}+c_{1} \times r_{1} \times \left(x_{t}^{\text {best }}-x_{t}\right)+ \\
&c_{2} \times r_{2} \times \left(x_{t}^{\text {gbest }}-x_{t}\right)
\end{aligned}
\end{equation}
\begin{equation}
x_{t+1}=x_{t}+v_{t+1}
\end{equation}
Where, $w$ is the inertial weight, $v_{t}$ and $v_{t+1}$ are velocity of the particle, related to the current and the next iteration, respectively. $c_1$ and $c_2$ are personal and global coefficients, respectively. $r_1$ and $r_2$ are random numbers between $0$ and $1$. $x_t^{gbest}$ is the best particle. $x_{t}$ and $x_{t}^{\text {best }}$  are the current position of the particle and the best position that it had experienced, respectively. Furthermore, velocity has lower and upper bounds. In this paper, maximum velocity is considered to be $10\%$ of the difference between upper and lower bounds of the optimized load. $c_1$ and $c_2$ are set as 2 , and $w$ is equal to $1$. Here, the stopping criteria is the iteration number, which is equal to $100$. The output will be load shifting pattern for residential customers of an area in Austin city.
\subsection{Differential Evolution Algorithm}
DE algorithm uses a unique method for generating new solutions, that distinguishes it from other evolutionary algorithms. It first generates an initial population, that all the members of this population are considered to be a solution. Next, a temporary solution using mutation operator is created, as shown in Eq.7.
\begin{equation}
y=\text{pop}_{a}+\beta \times \left(\text{pop}_{b}-\text{pop}_{c}\right)
\end{equation}
Where y is the temporary solution. $\text{pop}_{a}$, $\text{pop}_{b}$ and $\text{pop}_{c}$ are three random solutions, and $\beta$ is a scale factor in the range of $[0.2,0.8]$. Then a new solution with the help of the crossover operator will be created as presented in Eq. 8.
\begin{equation}
z_{j}= \begin{cases}y_{j} & r_{j} \leq P C R \text { or } j=j_{0} \\ x_{j} & \text { otherwise }\end{cases}
\end{equation}
Where each $x_{j}$ is a member of the initial population, and each $z_j$ is a new solution. The steps of the DE algorithm are as follows:
\begin{enumerate}
    \item Defining the parameters of the problem and algorithm such as population size, crossover probability, and objective function.
    \item Creating an initial population and evaluating it.
    \item Repeating below steps, until stop conditions are satisfied (maximum iteration).
    \begin{itemize}
        \item Creating an initial population and evaluating it.
        \item Using the crossover operator to create a new solution, then, evaluate it.
        \item The new solution will be selected if it is better than the current solution.
    \end{itemize}
    \item The output will be the best solution ever found, which is load shifting pattern and the favorable load curve.
\end{enumerate}
\section{Simulation Results}
The predicted and real electrical demand for December $28$, and September $20$, are shown in Fig. 1. Fig. 2 presents the outputs of train and test data. The corresponding correlation coefficient of the training and test data, that shows the linear dependence of the predicted and the real load, is equal to $0.9983$ and 0.9903, respectively. Fig. 3 depicts the training performance based on Mean Square Error (MSE). The best training performance is $0.00076$ at epoch $1000$. In addition, the value of MSE for the test data is $0.0043$.
\begin{figure}
\subfloat{\includegraphics[width = 3in]{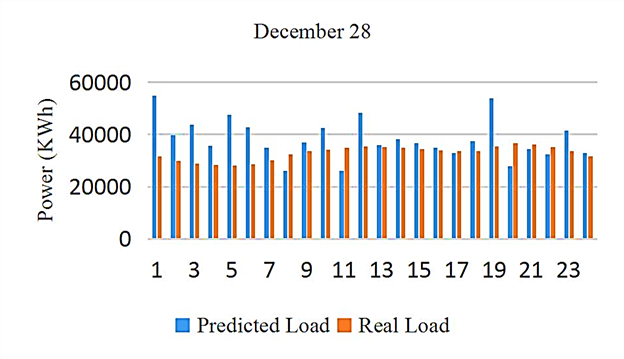}}\\
\subfloat{\includegraphics[width = 3in]{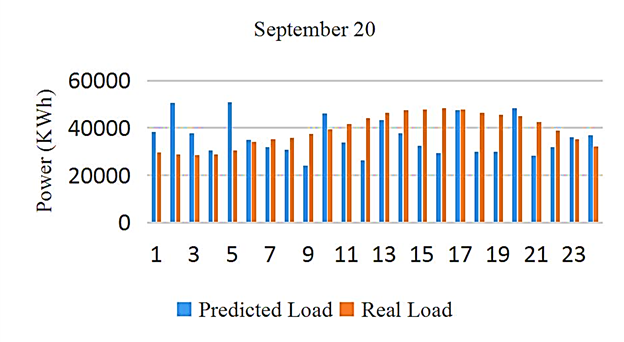}}
\caption{predicted vs real load for December 28 (top) and September 20 (bottom)}
\label{some example}
\end{figure}
\begin{figure}
\subfloat{\includegraphics[width = 3in]{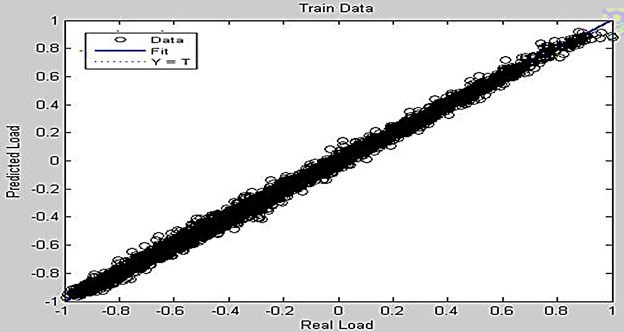}}\\
\subfloat{\includegraphics[width = 3in]{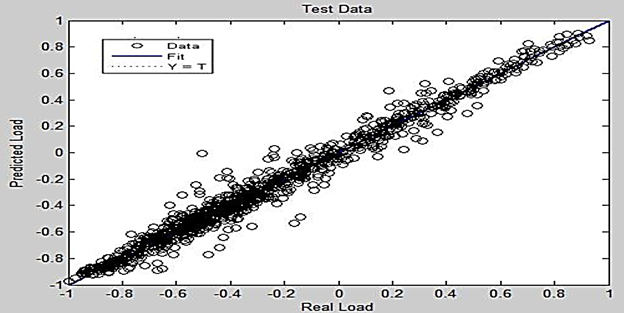}}
\caption{Train (top) and test (bottom) data results}
\label{some example}
\end{figure}
\begin{figure}[h]
    \centering
    \includegraphics[width=3in]{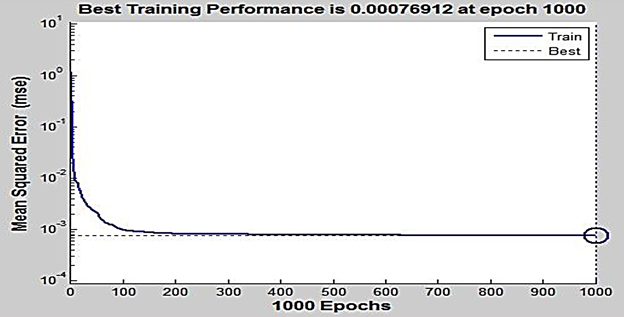}
    \caption{Training performance}
    \label{fig:my_label}
\end{figure}
\begin{figure}
\subfloat{\includegraphics[width = 3in]{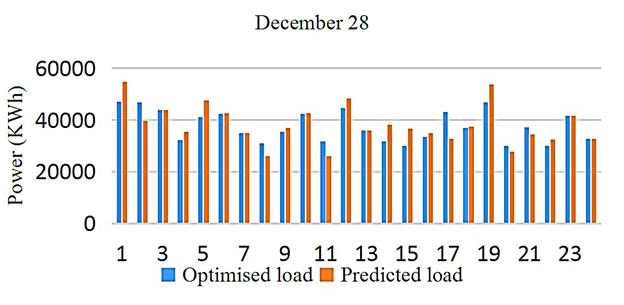}}\\
\subfloat{\includegraphics[width = 3in]{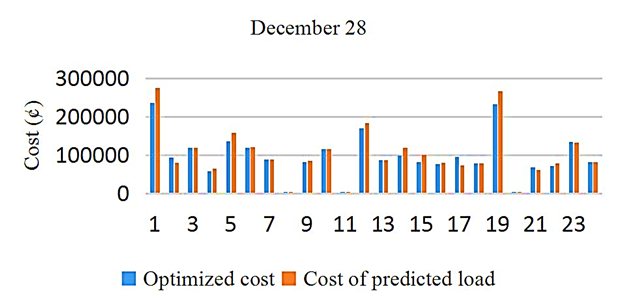}}
\caption{predicted vs optimized load (top) and cost (bottom) for December 28}
\label{some example}
\end{figure}
\begin{figure}
\subfloat{\includegraphics[width = 3in]{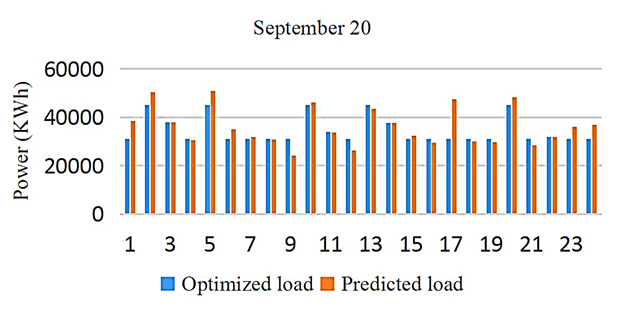}}\\
\subfloat{\includegraphics[width = 3in]{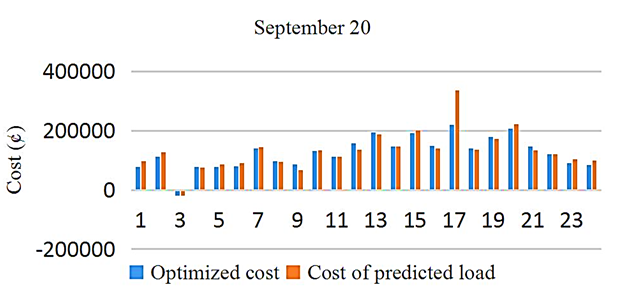}}
\caption{predicted vs optimized load (top) and cost (bottom) for September 20}
\label{some example}
\end{figure}
The results of PSO based DR load shifting for the selected days are provided in this section. Predicted load, optimized load, and corresponding costs are shown in Fig. 4 and 5. In regard to the results on December $28$, peak load is reduced from $54844.932KW$ to $47000KW$, and the daily cost is decreased from $24497.938\$$ to $23250.378\$$, leading to $5.09\%$ cost reduction. The results of September $20$, indicate that peak load is decreased from $50923.343KW$ to $45000KW$, and the cost is diminished from $31431.06\$$ to $29976.657\$$, which results in $4.62\%$ decrease in the costs.
Table I shows the effect of altering $w_1$ and $w_2$, on the total cost of energy, using the predicted load of May $5$. The cost of energy for the predicted load without using DR program is $31279.951\$$, which is more than the cost of all cases shown in Table 1. As illustrated, the optimized cost varies between $28598.554\$$ to $30621.802\$$, leading to a percentage of cost reduction of up to $8.57\%$, and maximum percentage of peak load reduction equal to $17.9\%$. Consequently, the best case is $w_1=0.4$ and $w_2=0.6$.
\begin{equation*}
\begin{aligned}
&\text { TABLE I. PSO based cost results on May } 5\\
&\begin{array}{|l|c|c|c|}
\hline W_{1} & W_{2} & \text { Cost (\$) } & \begin{array}{c}
\text { Percentage } \\
\text { of peak } \\
\text { load } \\
\text { reduction }
\end{array} \\
\hline 0 & 1 & 29917.812 & 17.73 \\
\hline 0.1 & 0.9 & 29975.347 & 17.9 \\
\hline 0.2 & 0.8 & 30621.802 & 17.75 \\
\hline 0.3 & 0.7 & 29737.152 & 17.73 \\
\hline 0.4 & 0.6 & 28598.554 & 17.9 \\
\hline 0.5 & 0.5 & 30482.678 & 17.73 \\
\hline 0.6 & 0.4 & 29818.165 & 17.73 \\
\hline 0.7 & 0.3 & 30350.973 & 17.75 \\
\hline 0.8 & 0.2 & 29270.375 & 17.73 \\
\hline 0.9 & 0.1 & 29855.073 & 17.73 \\
\hline 1 & 0 & 30494.192 & 17.73 \\
\hline
\end{array}
\end{aligned}
\end{equation*}
In order to assess the efficiency of the proposed model, the results of the PSO algorithm corresponding to July 10 , are compared with the results of the DE algorithm using the same data. Table II provides the results of these two algorithms for operating cost reduction and percentage of peak load reduction. Peak load of the predicted load is $60511.469KW$ and the corresponding cost of energy is $31389.529\$$. Both algorithms reduce cost and peak load. However, PSO shows better results in cost and peak load reduction.
\begin{equation*}
\begin{aligned}
&\text { TABLE II. Peak load and cost reduction results on July } 10,2010\\
&\begin{array}{|c|c|c|c|}
\hline \text {Algorithm} & \text {Total cost (\$)} & \begin{array}{c}
\text {Percentage of} \\
\text {cost reduction}
\end{array} & \begin{array}{c}
\text {Percentage of} \\
\text {peak load} \\
\text {reduction}
\end{array} \\
\hline \text {PSO} & 28929.995 & 7.83  & 22.32 \\
\hline \mathrm{DE} & 30723.182 & 2.12  & 15.75 \\
\hline
\end{array}
\end{aligned}
\end{equation*}
\section{Conclusion}
In this paper, a multilayer feed-forward ANN model is employed for prediction of the day-ahead load and the outcomes show a near correlation with the real load with a low value of MSE. Then, PSO algorithm is used to solve the problem of day-ahead load shifting and cost saving with DR program. The proposed model, as it is illustrated with the simulation results, is effective in improving the load curve and reducing the operating costs. The percentage of cost reduction is between $4.62\%$ to $8.57\%$. Moreover, peak load is decreased
up to $17.9\%$. Finally, to evaluate the performance of the PSO algorithm the results related to a specified day, are compared with the DE algonithm results, which shows the superiority of PSO.

\bibliographystyle{unsrt}
\bibliography{ref.bib}

\begin{thebibliography}{10}

\bibitem{kerdphol2016optimum}
Thongchart Kerdphol, Yaser Qudaih, and Yasunori Mitani.
\newblock Optimum battery energy storage system using pso considering dynamic
  demand response for microgrids.
\newblock {\em International Journal of Electrical Power \& Energy Systems},
  83:58--66, 2016.

\bibitem{6861959}
John~S. Vardakas, Nizar Zorba, and Christos~V. Verikoukis.
\newblock A survey on demand response programs in smart grids: Pricing methods
  and optimization algorithms.
\newblock {\em IEEE Communications Surveys Tutorials}, 17(1):152--178, 2015.

\bibitem{8681422}
Renzhi Lu, Seung~Ho Hong, and Mengmeng Yu.
\newblock Demand response for home energy management using reinforcement
  learning and artificial neural network.
\newblock {\em IEEE Transactions on Smart Grid}, 10(6):6629--6639, 2019.

\bibitem{haider2016residential}
Haider~Tarish Haider, Ong~Hang See, and Wilfried Elmenreich.
\newblock Residential demand response scheme based on adaptive consumption
  level pricing.
\newblock {\em Energy}, 113:301--308, 2016.

\bibitem{8385345}
Sukhlal Sisodiya, G.~B. Kumbhar, and M.~N. Alam.
\newblock A home energy management incorporating energy storage systems with
  utility under demand response using pso.
\newblock In {\em 2018 IEEMA Engineer Infinite Conference (eTechNxT)}, pages
  1--6, 2018.

\bibitem{del2008particle}
Yamille Del~Valle, Ganesh~Kumar Venayagamoorthy, Salman Mohagheghi, Jean-Carlos
  Hernandez, and Ronald~G Harley.
\newblock Particle swarm optimization: basic concepts, variants and
  applications in power systems.
\newblock {\em IEEE Transactions on evolutionary computation}, 12(2):171--195,
  2008.

\bibitem{kinhekar2015particle}
Nandkishor Kinhekar, Narayana~Prasad Padhy, and Hari~Om Gupta.
\newblock Particle swarm optimization based demand response for residential
  consumers.
\newblock In {\em 2015 IEEE power \& energy society general meeting}, pages
  1--5. IEEE, 2015.

\bibitem{faria2013modified}
Pedro Faria, Jo{\~a}o Soares, Zita Vale, Hugo Morais, and Tiago Sousa.
\newblock Modified particle swarm optimization applied to integrated demand
  response and dg resources scheduling.
\newblock {\em IEEE Transactions on smart grid}, 4(1):606--616, 2013.

\bibitem{raza2015review}
Muhammad~Qamar Raza and Abbas Khosravi.
\newblock A review on artificial intelligence based load demand forecasting
  techniques for smart grid and buildings.
\newblock {\em Renewable and Sustainable Energy Reviews}, 50:1352--1372, 2015.

\bibitem{chang2011monthly}
Pei-Chann Chang, Chin-Yuan Fan, and Jyun-Jie Lin.
\newblock Monthly electricity demand forecasting based on a weighted evolving
  fuzzy neural network approach.
\newblock {\em International Journal of Electrical Power \& Energy Systems},
  33(1):17--27, 2011.

\bibitem{baliyan2015review}
Arjun Baliyan, Kumar Gaurav, and Sudhansu~Kumar Mishra.
\newblock A review of short term load forecasting using artificial neural
  network models.
\newblock {\em Procedia Computer Science}, 48:121--125, 2015.

\bibitem{simon2004comprehensive}
Haykin Simon.
\newblock A comprehensive foundation.
\newblock {\em Neural networks}, 2(2004), 2004.

\bibitem{1527779}
Yu-Jun He, You-Chan Zhu, Jian-Cheng Gu, and Cheng-Qun Yin.
\newblock Similar day selecting based neural network model and its application
  in short-term load forecasting.
\newblock In {\em 2005 International Conference on Machine Learning and
  Cybernetics}, volume~8, pages 4760--4763 Vol. 8, 2005.

\bibitem{lee2009time}
Cheng-Ming Lee and Chia-Nan Ko.
\newblock Time series prediction using rbf neural networks with a nonlinear
  time-varying evolution pso algorithm.
\newblock {\em Neurocomputing}, 73(1-3):449--460, 2009.

\bibitem{crone2011advances}
Sven~F Crone, Michele Hibon, and Konstantinos Nikolopoulos.
\newblock Advances in forecasting with neural networks? empirical evidence from
  the nn3 competition on time series prediction.
\newblock {\em International Journal of forecasting}, 27(3):635--660, 2011.

\bibitem{gheyas2011novel}
Iffat~A Gheyas and Leslie~S Smith.
\newblock A novel neural network ensemble architecture for time series
  forecasting.
\newblock {\em Neurocomputing}, 74(18):3855--3864, 2011.

\bibitem{Electric}
Electric Reliability~Council of~Texas.
\newblock 2010.

\bibitem{Electric1}
National~Centre for Environmental~Information.
\newblock 2010.

\bibitem{Electric3}
ComEd’s Hourly~Pricing Program.
\newblock 2010.

\bibitem{gupta2021critical}
Varun Gupta, Monika Mittal, Vikas Mittal, and Nitin~Kumar Saxena.
\newblock A critical review of feature extraction techniques for ecg signal
  analysis.
\newblock {\em Journal of The Institution of Engineers (India): Series B},
  102(5):1049--1060, 2021.

\bibitem{hakak2021ensemble}
Saqib Hakak, Mamoun Alazab, Suleman Khan, Thippa~Reddy Gadekallu, Praveen
  Kumar~Reddy Maddikunta, and Wazir~Zada Khan.
\newblock An ensemble machine learning approach through effective feature
  extraction to classify fake news.
\newblock {\em Future Generation Computer Systems}, 117:47--58, 2021.

\bibitem{bayat2022human}
Nasrin Bayat, Elham Rastegari, and Qifeng Li.
\newblock Human gait recognition using bag of words feature representation
  method.
\newblock {\em arXiv preprint arXiv:2203.13317}, 2022.

\bibitem{kampelis2018development}
Nikos Kampelis, Elisavet Tsekeri, Dionysia Kolokotsa, Kostas Kalaitzakis,
  Daniela Isidori, and Cristina Cristalli.
\newblock Development of demand response energy management optimization at
  building and district levels using genetic algorithm and artificial neural
  network modelling power predictions.
\newblock {\em Energies}, 11(11):3012, 2018.

\end{thebibliography}

\end{document}